\documentclass[journal]{IEEEtran}
\pdfoutput=1

\usepackage{cite}

\usepackage[pdftex]{graphicx}
\DeclareGraphicsExtensions{.pdf,.jpeg,.png}

\usepackage{amsmath}
\usepackage{amssymb}
\usepackage{url}
\usepackage{hyperref}

\usepackage{color}

\newcommand{\repoLink}[1]{\textcolor{magenta}{#1}}

\usepackage{booktabs}
\usepackage{subfigure}
\usepackage{multirow}
\usepackage[export]{adjustbox}

\usepackage{arydshln}
\begin{document}
%
\title{Content-Aware Detection of\\ Temporal Metadata Manipulation}
%
%
%

\author{Rafael~Padilha, 
        Tawfiq~Salem, 
        Scott~Workman, 
        Fernanda~A.~Andaló, 
        Anderson~Rocha, 
        and~Nathan~Jacobs
        
\thanks{Corresponding author: \textbf{R. Padilha}, \textit{rafael.padilha@ic.unicamp.br}}%
\thanks{\textbf{R. Padilha}, \textbf{F. A. Andaló}, and \textbf{A. Rocha} are with the Institute of Computing, 
University of Campinas, Brazil.}
\thanks{\textbf{T. Salem} is with the Department of Computer and Information Technology, Purdue University, USA.}
\thanks{\textbf{S. Workman} is with DZYNE Technologies, USA.}
\thanks{\textbf{N. Jacobs} is with the Department of Computer Science, University of Kentucky, USA.}
\thanks{This paper has supplementary downloadable material available at \href{http://ieeexplore.ieee.org}{http://ieeexplore.ieee.org}, provided by the author. The material includes a list and description of the transient attributes and additional experiments highlighting aspects of our method. Contact \textit{rafael.padilha@ic.unicamp.br} for further questions about this work.}}

%
%

\markboth{IEEE Transactions on Information Forensics and Security, Vol. X, 2021}%
{Padilha \MakeLowercase{\textit{et al.}}: Content-Aware Detection of Temporal Metadata Manipulation}
%



\maketitle

\begin{abstract}
Most pictures shared online are accompanied by temporal metadata (i.e., the day and time they were taken),  which makes it possible to associate an image content with real-world events. Maliciously manipulating this metadata can convey a distorted version of reality. In this work, we present the emerging problem of detecting timestamp manipulation. We propose an end-to-end approach to verify whether the purported time of capture of an outdoor image is consistent with its content and geographic location. We consider manipulations done in the hour and/or month of capture of a photograph. The central idea is the use of supervised consistency verification, in which we predict the probability that the image content, capture time, and geographical location are consistent. We also include a pair of auxiliary tasks, which can be used to explain the network decision. Our approach improves upon previous work on a large benchmark dataset, increasing the classification accuracy from $59.0\%$ to $81.1\%$. We perform an ablation study that highlights the importance of various components of the method, showing what types of tampering are detectable using our approach. Finally, we demonstrate how the proposed method can be employed to estimate a possible time-of-capture in scenarios in which the timestamp is missing from the metadata. 
\end{abstract}


%
\IEEEpeerreviewmaketitle

\section{Introduction}
    \IEEEPARstart{W}{ith} the popularization of social networks and advances in image capturing devices during the last decade, the number of images shared online has grown exponentially. Their increasing availability coupled with easy-to-use photo editing software has resulted in a profusion of manipulated images. Such images are often maliciously used to support false claims and opinions. Consequently, the research community has explored many approaches to detect image content tampering~\cite{christlein2012evaluation, zhou2018learning, huh2018fighting}. Even though visual manipulation is now well-understood and reasonably explored in the digital forensics literature, there is a more subtle type of manipulation that still has been relatively unexplored by the research community: \emph{timestamp manipulation}. 
    
    Presenting an image as if it was captured in a different moment in time can corroborate a false narrative and spread misinformation. Recent examples include the ``\emph{Fishwrap}'' campaign~\cite{guardian2019fishwrap}, an online campaign that shared terror news articles from past years in an attempt to spread fear and uncertainty. In cases like this, articles often contain unaltered pictures that are displaced in time to validate a false story. Even though the timestamp of an image might be stored in its metadata when the photo was originally taken, it can be easily tampered with by freely available metadata editors (e.g., EXIF Date Changer Lite or Metadata++).

    \begin{figure}[t!]
        \centering
        \includegraphics[width=\linewidth]{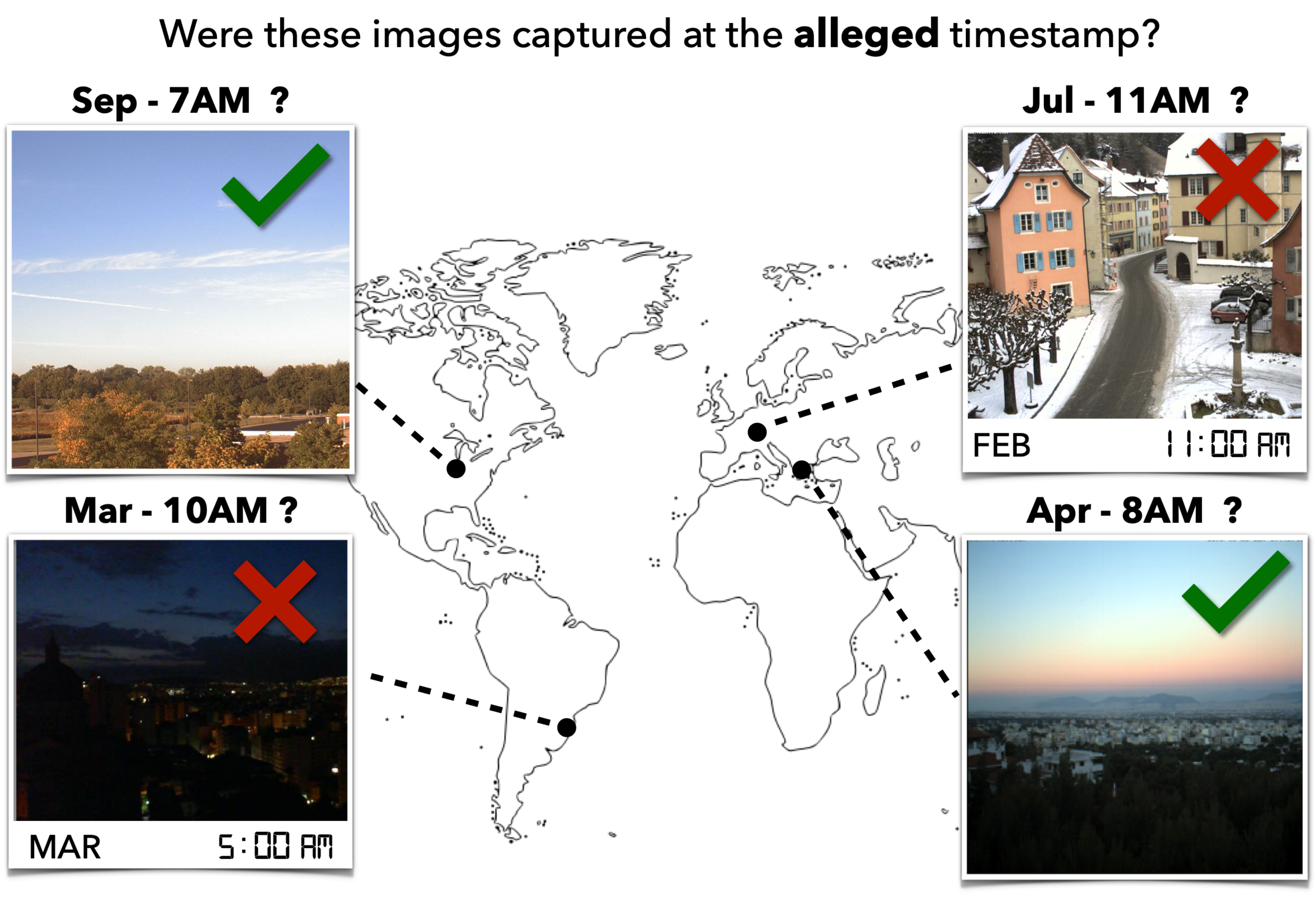}
        \caption{Our goal is to analyze if an \textbf{alleged} timestamp is consistent with the visual information present in a photograph. It is essential to consider the appearance of the scene (e.g., illumination and weather conditions) and photograph's geolocation when verifying a timestamp.}
        \label{fig:timestamp_verification}
    \end{figure}
    
    In this sense, it is important to develop methods capable of verifying the consistency between the visual information of a picture and its timestamp. This is a challenging task that requires a deep understanding of the scene, as its appearance may vary depending on the hour, month, and the location where the photo was taken (Figure~\ref{fig:timestamp_verification}). Moreover, factors such as the weather, lighting conditions, device quality, and depicted elements influence the appearance of a recorded scene and directly affect our perception of time. Existing methods often estimate indirect features from the image content, such as the sun position in the sky~\cite{ghosh2017detection, kakar2012verifying, li2017you} or meteorological measures~\cite{ghosh2017detection}, and contrast them to registered values for the same day, hour, and location. However, these are limited cues that may not always be sufficient nor easily available.
    
    We propose a convolutional neural network (CNN)-based solution that analyzes a ground-level outdoor image and a claimed timestamp (hour and month), receiving geographical information as an additional context for its decision. The geo-information can take the form of location coordinates (i.e., latitude and longitude) and/or basemap-style satellite imagery associated with the location coordinates (but not necessarily with the claimed timestamp). Our model is optimized in an end-to-end manner using a multi-task loss function, with the main goal of checking for manipulations and an auxiliary objective of estimating transient attributes~\cite{laffont2014transient}. These are a set of 40 attributes that encode the presence of high-level properties of the scene appearance, such as weather (\emph{snow}, \emph{rainy}, \emph{sunny}), period of the day (\emph{sunrise}, \emph{night}, \emph{dusk}), and even subjective concepts (\emph{beautiful}, \emph{stressful}). This secondary task introduces an explainability component to our model, aiding in understanding its decisions. 
    
    We evaluate the proposed approach both quantitatively and qualitatively, achieving state-of-the-art results on a reference benchmark dataset~\cite{salem2020learning}. Moreover, we perform sensitivity analyses to understand how the appearance of the scene, subtler timestamp manipulations, and noisy location coordinates affect the verification performance of the method. \\

    \noindent\textbf{The contributions of our work include:}
    \begin{itemize}
        \item A new method for verifying the hour and month of capture of an outdoor image by comparing the appearance of the scene with an alleged time and location information.
        \item An extension of our method to estimate a possible time-of-capture in scenarios in which the timestamp is missing from the metadata.
        \item A high-level exploration of which elements in a scene might indicate timestamp inconsistency, as well as several sensitivity analyses to help us understand when the method works and fails.
        \item A novel organization of the Cross-View Time dataset~\cite{salem2020learning} with camera-disjoint training and testing sets that are more closely related to real-world scenarios.
    \end{itemize}

\section{Related Work}~\label{sec:related_work}
    The analysis of temporal information has been explored in different ways in the literature. In this section, we review relevant methods that approach this problem. 

    \subsection{Metadata Tampering Detection}
        Traditionally, most tampering detection techniques focus on image content manipulation, with only a few recent works approaching metadata tampering detection. To check the integrity of the geographic location, researchers borrow ideas extensively from the image retrieval and landmark recognition literature~\cite{hays2008im2gps, noh2017large, radenovic2018fine, weyand2016planet}, focusing on identifying visual elements in the scene and matching them with large-scale databases of geo-tagged images.
        
        Considering timestamp manipulation, Kakar et al.~\cite{kakar2012verifying} and Li et al.~\cite{li2017you} propose to verify the timestamp by estimating the sun azimuth angle from shadow angles and sky appearance, comparing it to the sun position calculated from the image metadata. Their methods assume the camera to be perpendicular to the ground and that the sky and at least one shadow from vertical structures are visible in the scene. Chen et al.~\cite{ghosh2017detection} optimize a CNN to jointly estimate temperature, humidity, sun altitude angle, and weather condition from an input image, comparing them with meteorological data registered from the day and time stored in its metadata. Even though they achieve promising results, their approach requires access to historical weather data that might not be available for most locations. 
        
        In~\cite{chen2019deep}, the authors approach this task as an event identification problem: they train a CNN with images from a specific event (i.e., sharing similar location and timestamp) to determine whether a test image, presumably sharing the same metadata, belongs to that event. This assumption might be valid considering forensic events with high media coverage, but it might not be the case when most images are collected from social media. 
        
        Differently from the mentioned works, we aim at directly verifying whether a claimed time-of-capture is \emph{consistent} with the visual content of an outdoor image, independent of particular visual cues and without the need to optimize for a specific event. With this in mind, we do not hold strong assumptions on how the image was captured nor the appearance of the scene.   
        
        In a similar line, the method proposed by Salem et al.~\cite{salem2020learning} learns a dynamic map of visual attributes that captures the relationship between location, time, and the expected appearance of a photograph. Their approach predicts visual attributes from a combination of satellite imagery, time, and geographic location. To apply the method to metadata verification, these attributes are compared to similar information extracted from a ground-level picture, computing a distance metric and using it as a consistency score to detect if a timestamp has been tampered with. In a similar manner, we train a global model that captures how visual appearance relates to location and time. However, we specifically focus on contrasting an alleged timestamp against the visual characteristics of a picture by directly optimizing for this task.

    \subsection{Time-of-capture Estimation}
        Approaching the problem from a different perspective, we also analyze methods for directly estimating time-of-capture, with time scales ranging from hours to decades. Several works leverage specific visual elements as cues to date pictures, such as human appearance and fashion~\cite{ginosar2015century, salem2016analyzing}, visual style of objects~\cite{jae2013style, vittayakorn2017made}, architecture styles~\cite{linking2015iccp}, sun position~\cite{kakar2012verifying, li2017you, tsai2016photo}, and photo-generation artifacts~\cite{fernando2014color, martin2014dating, palermo2012dating}.
        
        Despite their impressive results, these methods require the presence of particular visual elements in a scene to estimate time reliably. In their absence, a different class of methods should be considered. In this sense, a line of research closely related to this work explores how the global appearance of a scene changes over time. 
        
        Volokitin et al.~\cite{volokitin2016deep} optimize a scene-specific classifier on top of data-driven features to infer the time of the year and hour of the day of images from a particular location. Some works~\cite{baltenberger16transient, laffont2014transient} aim to estimate transient attributes of a scene that inherently carry some degree of temporal information, such as season and illumination conditions. By combining the visual content with geographical information in the pretext task of time-of-capture estimation, Zhai et al.~\cite{zhai2018geotemporal} learn a representation with high correlation to transient attributes.
        
        This work builds upon these strategies, incorporating visual information from a ground-level photo, geographical coordinates to check for time-of-capture consistency, and optional satellite imagery. Even though our goal is not to explicitly estimate when an image was taken, we show how our method can be applied for cases in which the timestamp is missing.

    \begin{figure*}
        \centering
        \includegraphics[width=0.8\linewidth]{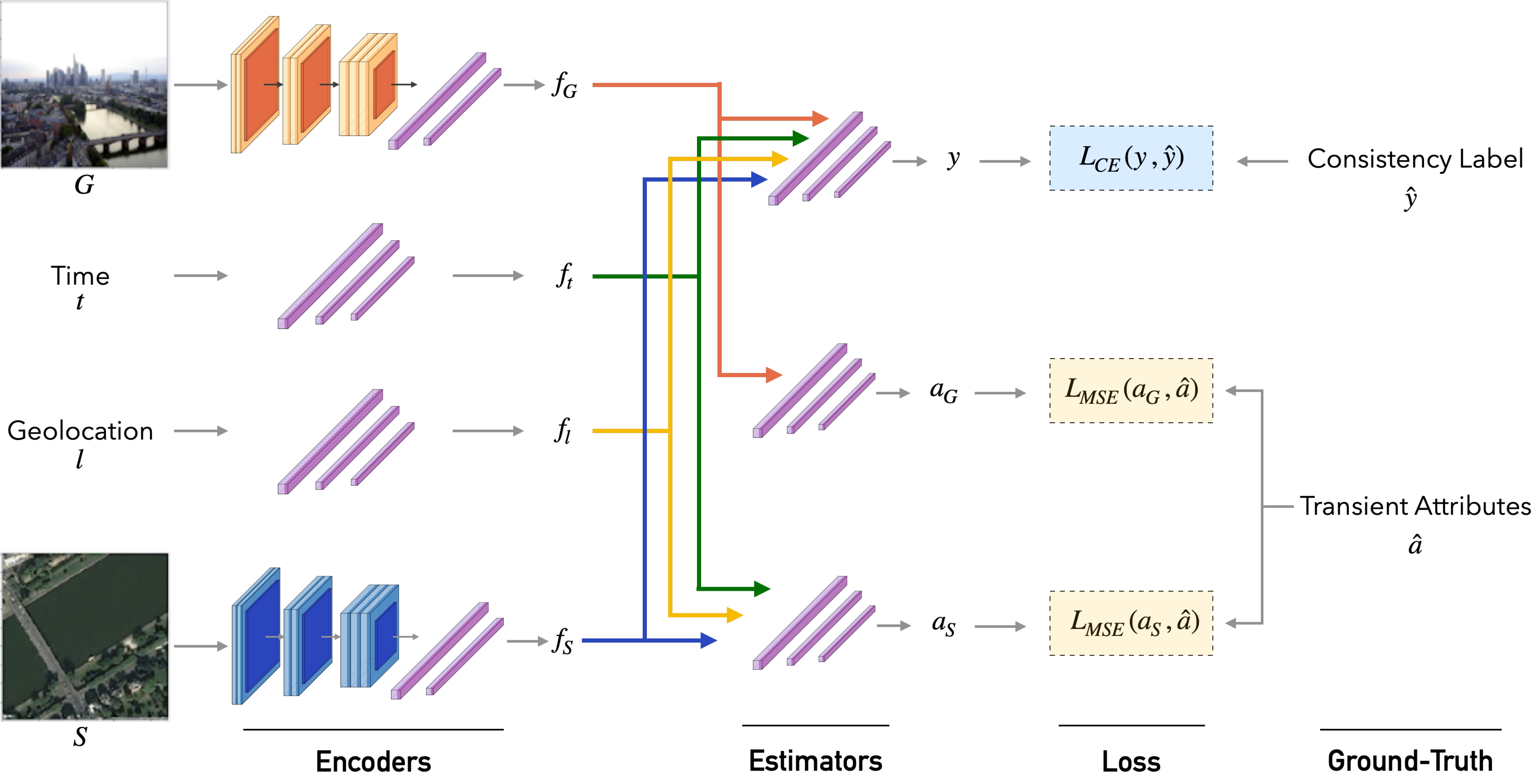}
        \caption{An overview of our approach. A ground-level image $G$, a timestamp $t$, geo-coordinates $l$, and a satellite image $S$ have their features extracted by encoder networks and then fed to task-specific branches. One network branch predicts the consistency label $y$, while auxiliary branches estimate transient attributes $a_{G}$ and $a_{S}$. The network is optimized by a combination of cross-entropy and mean squared error losses.  At inference time, the consistency answer $y$ is considered for the tampering detection, while $a_{G}$ and $a_{S}$ offer insights about the decision of the network.}
        \label{fig:architecture}
    \end{figure*}

\section{Proposed Method}
    Our goal is to assess if the visual content of an outdoor image is consistent with its hour and month of capture. For this, our method must extract discriminative features from the scene appearance and contrast them with the expected appearance for that specific time of capture. As variations in appearance over time are highly dependent on the location of the scene, it is essential to provide, as additional context, geographic cues of where the picture was taken~\cite{salem2020learning, zhai2018geotemporal}.  
    
    With this in mind, we propose a CNN architecture (Figure~\ref{fig:architecture}) to estimate the probability $P(y\,|\,G, t, l, S)$ that a given ground-level image $G$, associated with location $l$ and satellite image $S$, is consistent ($y=0$) or inconsistent ($y=1$) with an alleged timestamp $t$. By providing location $l$ as input, the network will be able to consider the influence of geographic position in seasonal patterns (e.g., winter months in the Northern hemisphere with reduced sunlight hours, and the opposite in the Southern hemisphere). Moreover,  satellite image $S$ provides an additional context about the photographer's surroundings and the structure of the scene, such as whether the image was captured in an urban or rural area. We employ a basemap-style picture, which is globally available and can be easily obtained from online services (e.g., Google Maps and Bing Maps) given location $l$. This kind of imagery was designed for navigational purposes and offers an idea of the structure of the scene without reflecting time-dependent elements (e.g., illumination and weather conditions). In this sense, we \textbf{do not} assume $S$ is linked to the timestamp $t$, avoiding the need for a satellite image at the precise time-of-capture being checked.
    
    For explainability, the network also estimates transient attributes $a_{\textit{G}}$ and $a_{\textit{S}}$. These are 40-dimensional arrays, with each value encoding the presence of a characteristic of the scene appearance (e.g., fog, hot, beautiful, summer) to the interval $[0,1]$. The attributes $a_{\textit{G}}$ are estimated solely from the ground-level image and capture the high-level properties of the scene at the moment it was recorded. In contrast, $a_{S}$ is estimated from the satellite photo, location coordinates, and the alleged timestamp, and can be interpreted as a prediction of the expected scene appearance at the alleged moment.
    
    \subsection{Network Architecture}
        Each input $G, t, l, S$ is processed by individual sub-networks that extract characteristics from each modality and encode them into 128-dimensional feature vectors. \newline
        
        \noindent\textbf{Visual Encoder.} Both the ground-level and satellite images, $G$ and $S$, are processed by a backbone CNN, extracting feature maps from the last convolutional layer. The feature maps are then processed by two additional fully-connected layers, with 256 and 128 units, respectively, each followed by a ReLU activation and batch normalization, resulting in feature vectors $f_{\textit{G}}$ and $f_{\textit{S}}$. \newline
        
        \noindent\textbf{Location and Time Encoder.} We represent location $l$ in earth-centered earth-fixed (ECEF) coordinates, scaled to $[-1, 1]$ by dividing each one by the radius of the Earth; whereas time $t$ is represented by the month and hour of the day (UTC) individually scaled to $[-1, 1]$ (i.e., 0 AM of January would lead to $(-1, -1)$ and 11 PM of December to $(1, 1)$ coordinates). Both sub-networks have similar architectures and comprise three fully-connected layers, with 256, 512, and 128 neurons, respectively, each followed by a ReLU activation function and batch normalization, resulting in feature vectors $f_{l}$ and $f_{t}$. \newline
        
        \noindent\textbf{Task-specific Branches.} Once each feature $f_{\textit{G}}$, $f_{\textit{S}}$, $f_{l}$, and $f_{t}$ has been extracted, they are fed to task-specific branches. Each branch consists of three fully-connected layers. The initial two layers have 256 and 512 units, respectively, followed by ReLU and batch normalization, while the third maps the intermediate features to the output dimension of each task. The top branch receives the concatenation of $\{f_{\textit{G}}$, $f_{\textit{S}}$, $f_{l}$, $f_{t}\}$ and outputs the consistency $y$. Its last fully-connected layer has two units followed by a softmax operation, allowing us to interpret the output as the probability $P(y\,|\,G, t, l, S)$. The middle branch receives as input $f_{\textit{G}}$ and outputs a set of transient attributes $a_{\textit{G}}$. Likewise, the bottom branch processes the concatenation of $\{f_{\textit{S}}, f_{l}, f_{t}\}$ and outputs attributes $a_{\textit{S}}$. Their final dense layers have a number of neurons matching the quantity of estimated transient attributes $|\hat{a}|$ (in our experiments, $|\hat{a}| = 40$), followed by a sigmoid activation.  
    
    \subsection{Loss Function}
        Our network is optimized in an end-to-end fashion both for timestamp consistency and transient attribute estimation tasks. Even though the main goal is timestamp tampering detection, the auxiliary tasks allow our model to predict properties of the scene, separately considering the alleged timestamp or the ground-level visual information. Both sets of attributes can be compared, offering insights into the model decision. 
        
        For the consistency verification branch, we calculate the binary cross-entropy loss
        \begin{equation}
            L_{\textsc{ce}}(y, \hat{y}) = -\hat{y}\log(y) - (1-\hat{y})\log(1-y)\; 
        \end{equation}
        
        \noindent between the consistency prediction $y$ and the ground truth $\hat{y}$. For the transient attribute branches, we compute the mean squared error between estimated transient attributes $a$ and ground truth $\hat{a}$:
        \begin{equation}\label{eq:mse}
            L_{\textsc{mse}}(a, \hat{a}) = \frac{1}{|\hat{a}|}\sum_{i=1}^{|\hat{a}|}(a_i - \hat{a_{i}})^2  \;.
        \end{equation}
        
        \noindent We compute two separate $L_{\textsc{mse}}$ terms with respect to $a_{G}$ and $a_{S}$. By doing so, the model learns to extract the transient attributes directly from the ground-level image while also being able to estimate them from the satellite photo, location coordinates, and given timestamp. Finally, the whole network is jointly optimized to minimize their weighted sum:
        \begin{equation}\label{eq:loss}
            L = \alpha L_{\textsc{ce}}(y, \hat{y}) + \beta L_{\textsc{mse}}(a_{G}, \hat{a}) + \gamma L_{\textsc{mse}}(a_{S}, \hat{a}) \;.
        \end{equation}

        \noindent Even though setting different weights ($\alpha$, $\beta$ and $\gamma$) to each loss term in Equation~(\ref{eq:loss}) might allow the network to focus on a particular task and improve the detection performance, in our experiments, we found that equal importance to all terms achieved better results. We perform our experimental analysis in Section~\ref{sec:exp_analysis} with this design decision.

\section{Experimental Analysis}\label{sec:exp_analysis}
    We evaluated our method for tampering detection, in which it outputs whether the timestamp is consistent with the visual attributes of the ground-level image. We assessed the quality of our approach considering the accuracy (Acc) and the receiver operating characteristic (ROC) curves in the tampering detection. We performed an ablation study of different input modalities and the backbone CNNs used as visual encoders, comparing them to existing approaches from the literature. We also evaluated its sensitivity to realistic conditions, with changes in the appearance of the scene, subtler timestamp manipulations, and noisy location information. Additionally, for scenarios in which the timestamp of an image is missing, we demonstrated how our method could be applied to estimate a possible moment of capture. Finally, we interpret the decisions of our approach by analyzing image regions with strong influence on the model's predictions and mismatches between estimated transient attributes.

    \subsection{Dataset and Training Details}~\label{sec:dataset_training_details}
        We adopted the Cross-View Time dataset~\cite{salem2020learning}, comprising more than 300k ground-level outdoor images. The dataset combines 98k images from 50 static outdoor webcams of the Archive of Many Outdoor Scenes (AMOS)~\cite{jacobs2007consistent}, and 206k geo-tagged smartphone pictures from the Yahoo Flickr Creative Commons 100 Million Dataset~\cite{thomee2016yfcc100m}.  Ground-level images were captured worldwide in different hours of the day and months throughout the year, and are associated with geographical coordinates and a timestamp (UTC). Given the latitude and longitude of ground-level pictures, the authors also collected co-located satellite imagery, downloading from Bing Maps orthorectified overhead image centered on the geographic location. Images have dimensions of $800\times800$ pixels and a spatial resolution of 0.60 meters/pixel. These are not captured at the same moment of the timestamp, but instead are basemap-style imagery intended to provide a clear picture of the current configuration of buildings and roads. This type of photograph offers plenty of details about the surrounding of the photographer without reflecting weather conditions at the time of capture (e.g., clouds, snow, rain). Lastly, the authors provided for each ground-level image the set of 40 transient attributes extracted with the method from~\cite{laffont2014transient} and encoded to the interval $[0, 1]$.

        We employed the same data splits provided by the authors, training on 280k images and testing on 25k. The original splits share locations (i.e., the same place might be represented through different images in training and testing) and cameras, as imagery from AMOS has been randomly sampled for each set. In Section~\ref{sec:cross_camera}, we propose a novel and challenging organization of the dataset that considers camera-disjoint sets.
        
        For training, batches were randomly sampled from the training images, with corresponding timestamps, geo-coordinates, and satellite images. For each consistent sample, we generated a tampered version by exchanging its timestamp to that of a random image in the training set. Due to the hour distribution of the set --- which mostly comes from social media and, thus, is concentrated on the 10 AM to 8 PM interval in which people take more photos --- it is less likely that the majority of images will completely change their period of the day after the manipulation. This type of tampering creates a more realistic scenario for our evaluation than randomly sampling a tampered timestamp throughout the day and year, as done by previous works~\cite{li2017you}. Considering this, training batches were composed of the same number of consistent and inconsistent tuples. At test time, a similar process was used, generating a tampered tuple for each available test image.\footnote{The same seed was selected for all experiments and the tampered timestamps of test images are available in \href{https://github.com/rafaspadilha/timestampVerificationTIFS}{\repoLink{https://github.com/rafaspadilha/timestampVerificationTIFS}}}
        
        We use a VGG-16~\cite{simonyan2014very} network pre-trained on the \emph{Places} dataset~\cite{zhou2017places} as the feature extractor for the ground-level image, while a ResNet-50~\cite{he2016identity} network pre-trained on \emph{ImageNet}~\cite{deng2009imagenet} processes the satellite image. This is a similar architecture to~\cite{salem2020learning} which allows a fair comparison to their method. Before being processed by the networks, we resize $G$ and $S$ to $224\times224$ and scale each pixel to $[-1, 1]$.
        
        Our architecture was optimized using Adam~\cite{kingma2014adam} with an initial learning rate of $10^{-5}$, batches of 32 images, and trained for 30 epochs. When calculating the loss for a batch, only real tuples were considered in computing the $L_{\textsc{mse}}$ terms of Equation~(\ref{eq:loss}), as the ground-truth transient attributes for tampered timestamps are not available. With the exception of VGG-16 and ResNet-50 sub-networks, the weights of convolutional and fully-connected layers were initialized with Xavier initialization, and we applied $L_2$ regularization ($\lambda = 0.001$).

    \subsection{Ablation Study}\label{sec:exp_ablation}
    
        Each input modality influences the performance of the timestamp tampering detection. We performed an ablation study, considering the impact of location $l$ and satellite image $S$, given as additional context to the ground-level image $G$ and timestamp $t$. To better evaluate the impact of each modality, we optimized the models considering solely the consistency verification, i.e., by removing the mean-squared error terms from Equation~(\ref{eq:loss}). We also considered variations of our architecture in which we use a ResNet-50~\cite{he2016identity} or a DenseNet-121~\cite{huang2017densely} as encoders for both ground-level and satellite images. In this scenario, even though the branches are similar, they do not share weights as each processes a different type of input. Finally, considering all input modalities and evaluated backbones, we optimized the model with the transient attribute estimation task by employing the complete loss function from Equation~(\ref{eq:loss}). We present the accuracy and the area under the receiver operating characteristic curves (AUC) for our evaluation in Table~\ref{tab:ablation}.
        
        \begin{table}[!t]
            \centering
            \caption{Results for the ablation study of input modalities and visual encoder architectures.} 
            \label{tab:ablation}
            \resizebox{\linewidth}{!}{%
            \begin{tabular}{@{}lcccccc@{}}
            \toprule
            &  \multicolumn{2}{l}{VGG-16 / ResNet-50} & \multicolumn{2}{c}{ResNet-50} & \multicolumn{2}{r}{DenseNet-121} \\ \cmidrule(lr){2-3} \cmidrule(lr){4-5} \cmidrule(l){6-7} 
            Modalities      &  Acc (\%)      & AUC     & Acc (\%)    & AUC  &  Acc (\%) & AUC\\ \midrule
            G, t           &  $63.6$  & $.699$ &   $65.3$ & $.744$   &   $67.5$ & $.766$\\
            G, t, l         &  $\textbf{77.9}$  & $\textbf{.853}$ &   $75.1$ & $.852$   &   $78.7$ & $.873$\\
            G, t, S         &  $72.1$  & $.813$ &   $74.9$ & $.851$   &   $77.0$ & $.855$\\
            G, t, l, S      &  $76.1$  & $.847$ &   $78.4$ & $\textbf{.877}$   &   $80.5$ & $.880$\\
            G, t, l, S (TA) & $75.7$  & $.834$ &   $\textbf{78.7}$ & $.865$   &   $\textbf{81.1}$ & $\textbf{.885}$\\\hdashline
            Salem et al.~\cite{salem2020learning} & $59.0$   & $.627$ &  -- & -- & -- & --\\ 
            Volokitin et al.~\cite{volokitin2016deep} & $51.7$   & $.517$ &  -- & -- & -- & --\\\bottomrule         
            \end{tabular}%
            }
        \end{table}
        

        Our results show that including geographical coordinates ($G, t, l$) considerably improved performance, as the model learns how shifts in latitude and longitude correlate to the expected appearance of a scene and an alleged timestamp (e.g., a snowy scene in December might be consistent if taken in the Northern hemisphere, but less so in the Southern hemisphere). The inclusion of satellite imagery ($G, t, S$) also boosts performance, but the gain is slightly inferior to that of the location-only model ($G, t, l$). In addition, when comparing to each individually ($G,t,S$ and $G,t,l$ models), the simultaneous inclusion of coordinates and satellite data improves the results even further ($G,t,l,S$). This indicates that both input modalities share complementary information that is being captured by the model. Even though one might think the aerial modality bears a collection overhead that might not be worth considering its performance gain, satellite data is globally available and can be directly obtained from location coordinates. In this sense, it would be more likely not to have any sort of geographical information (e.g., if the metadata is completely erased) than not being able to collect an aerial image. Nonetheless, the location-only model ($G, t, l$) still achieved competitive results and could be employed in real application scenarios as well.
        
        Replacing the backbone CNNs to ResNet-50 or DenseNet-121 improved the results even further. Even though VGG-16 is frequently used for transfer learning, the other architectures include several operations---such as residual learning and batch normalization---that boost optimization and performance.
        
        Finally, the addition of the auxiliary optimization tasks---$G$, $t$, $l$, $S$ (TA) in Table~\ref{tab:ablation}---slightly improved the results for the ResNet-50 and DenseNet-121 setups. Even though the estimated attributes, $a_G$ and $a_S$, are not considered when determining if tampering occurred, in Section~\ref{sec:xai} we use them to produce a simple explanation of the network's decision.
        
       \begin{figure*}[!t]
            \centering
            \subfigure[Same location recorded in April under different hours of the day.]{
                \includegraphics[width=0.48\textwidth]{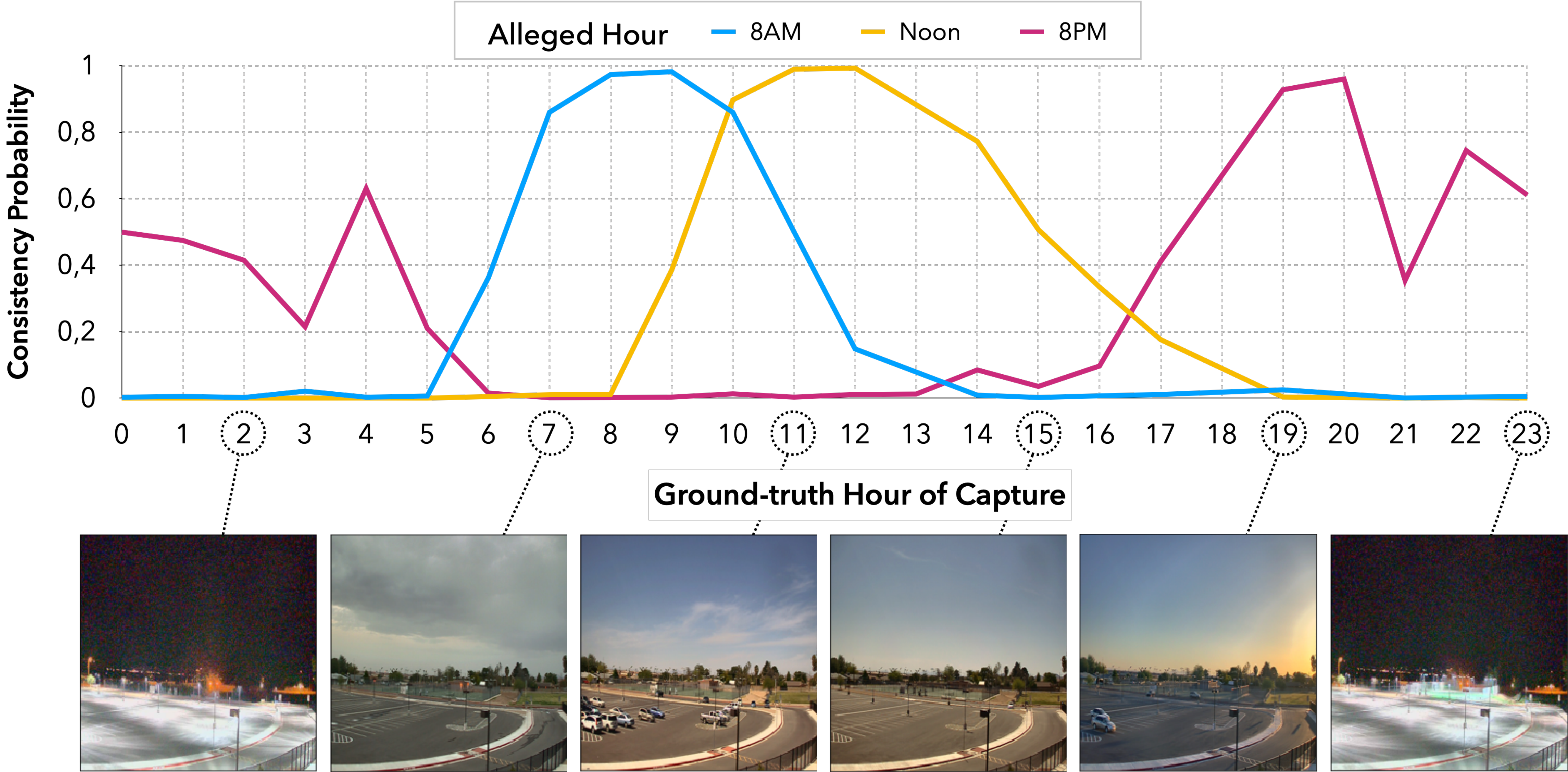}
                \label{fig:consistency_hour}
            }
            \hfill
            \subfigure[Same location recorded at 6\textsc{PM} in different months of the year.]{
                \includegraphics[width=0.48\textwidth]{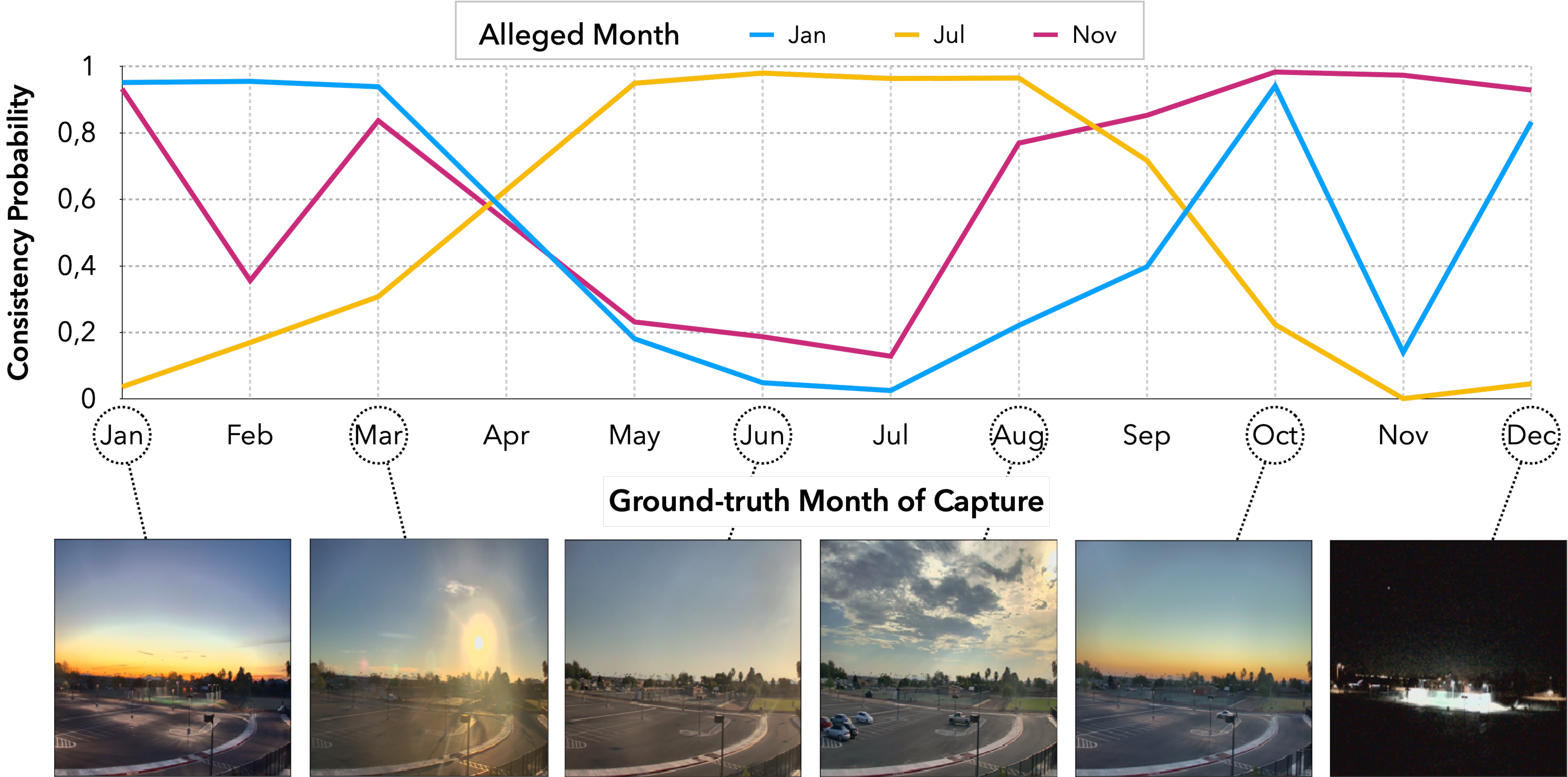}
                \label{fig:consistency_month}
            }
            
            \caption{Consistency probability for a scene recorded in different moments in time. Each curve represents a fixed alleged timestamp being verified against ground-level images captured in different \textbf{(a)} hours of the day in April and \textbf{(b)} months of the year at 6\textsc{PM}. Our model captures several temporal patterns, identifying how the appearance of the scene changes according to the period of the day and season.}
            \label{fig:consistency_variation}
        \end{figure*}
        
    \subsection{Comparison to Existing Approaches}\label{sec:exp_comparison_sota}        
        We compared our models with the method proposed by Salem et al.~\cite{salem2020learning}, using the same weights and choice of hyperparameters provided by the authors. They employ a similar architecture to ours---with VGG-16 and ResNet-50 as feature extractors for the ground-level and satellite imagery, respectively---that extracts features related to scene classification and transient attributes. In a similar fashion, one set of features is obtained directly from the ground-level picture and another from the satellite image, location, and timestamp. They compute a distance between both sets (based on KL divergence and $L_2$), using it as a consistency score for timestamp tampering detection. 
        
        Even though the authors apply their approach for timestamp verification, the learning process focuses on capturing the geographical and temporal behavior of transient attributes. Though this may aid in learning richer features, as shown by our ablation study, they are not discriminative enough by themselves for verifying time-of-capture. Our results show that optimizing the network specifically for this task was essential to better detect manipulations, outperforming their approach by $22$ percentage points in accuracy.
        
        Additionally, we evaluate the method of Volokitin et al.~\cite{volokitin2016deep} for this task. We trained Random Forest classifiers on top of features extracted with a pre-trained VGG model to estimate the month and hour of capture of an image. To apply their approach to the verification scenario, we computed the class probability of month and hour classifiers for each test image, multiplied them, and compared against a threshold selected in a held-out validation set.
        
        Differently from ours, their method does not leverage geographic information but instead trains each classifier with images from a single location. This might be viable in particular cases in which there are enough images to train a location-specific verification model but becomes infeasible as we analyze geographically-spread imagery, such as the CVT dataset. Moreover, optimizing for verification---instead of adapting a timestamp estimation method---allows our model to capture contrasting information between alleged timestamp and scene appearance essential to detect manipulations.
        
        Finally, most works mentioned in Section~\ref{sec:related_work} have strong assumptions about the input data that do not hold valid for images originating from social media. In such a scenario, available data is collected under unconstrained capturing conditions (i.e., presenting varied illumination and weather conditions, from different points of view and angles, having varied quality and resolution). Even though methods that rely on identifying particular visual clues (e.g., shadows from vertical structures, sun position in the sky, architectural or object style, fashion) might achieve accurate results when their underlying assumptions are met, they are often inadequate for general situations such as the one we tackle.
        

    \subsection{Sensitivity Analysis: Scene Appearance}\label{sec:sensitivity_image}
        We explored the sensitivity of our method to changes in the appearance of a scene under fixed alleged timestamps. In a first evaluation, we selected images from different hours of the day, all taken in a particular month at the same location. Whereas, in a second experiment, we employed images taken at the same hour and location but in different months. We computed the consistency probability of each image under fixed alleged timestamps, and plotted the resulting curves in Figure~\ref{fig:consistency_variation}. By doing so, we investigate how confident the network is that an alleged timestamp is consistent as time progresses in a scene.
        
        The model correctly predicts high probabilities for images taken around the alleged timestamp, as seen by the peaks in the curves from Figure~\ref{fig:consistency_hour}. Despite that, nighttime images tend to be very similar due to the lack of changes in illumination, reflecting in higher consistency probabilities between 9PM and 4AM. Similarly, Figure~\ref{fig:consistency_month} shows that our model captured seasonal patterns, such as the variation in sunset hours across the year, even though monthly variations in the appearance of a scene tend to be smoother across neighboring months. We present in the Supplementary Material additional examples and an evaluation considering multiple cameras from AMOS.

        \begin{figure*}
            \centering
            \subfigure[Model trained with random manipulations.]{
                \includegraphics[width=0.4\linewidth]{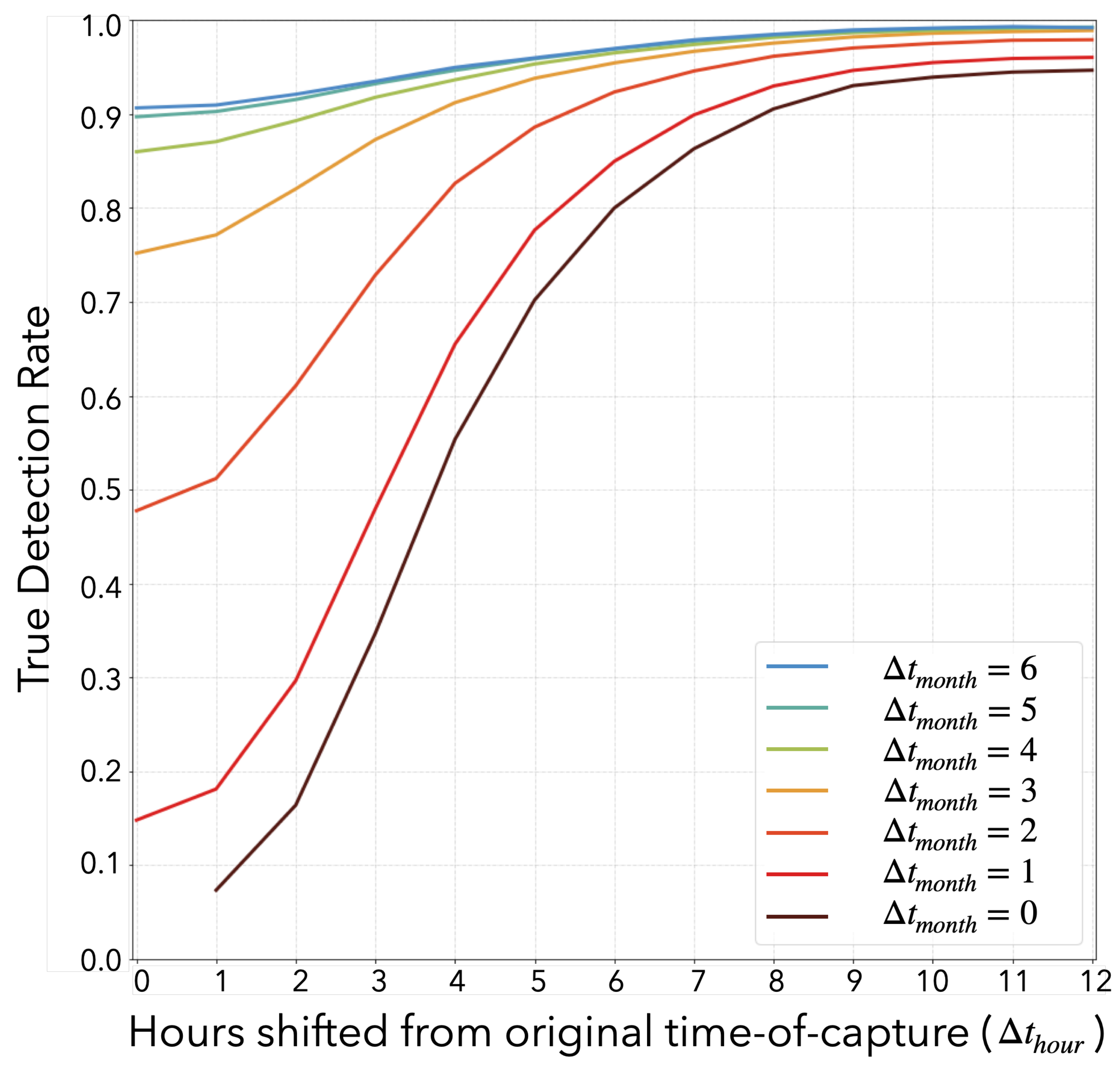}
                \label{fig:range_tampering_random}
            }
            \hfill
            \subfigure[Model fine-tuned for subtler manipulations.]{
                \includegraphics[width=0.4\linewidth]{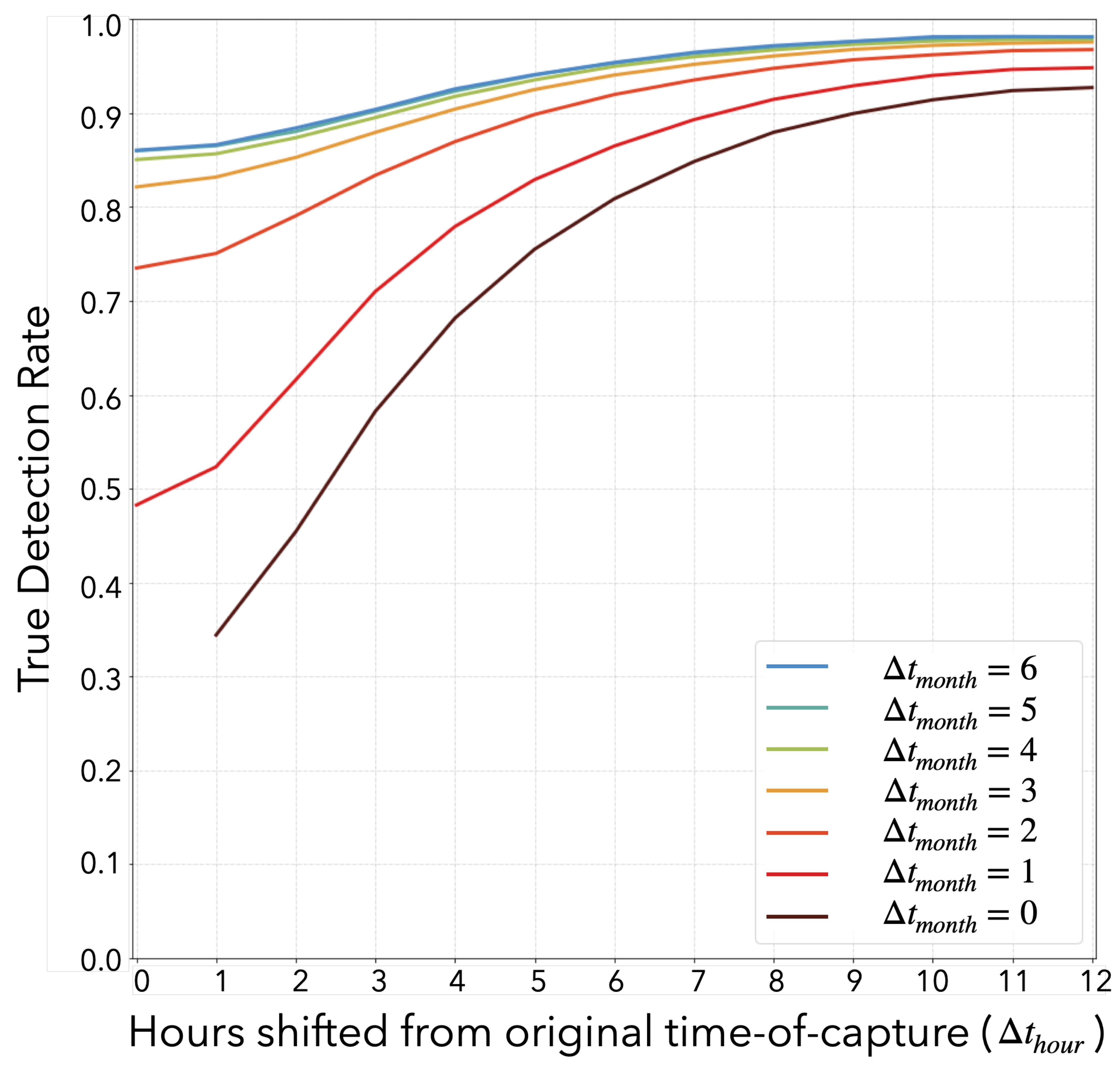}
                \label{fig:range_tampering_ngb}
            }
            
            \caption{Detection rate for hour and month shifts from the original timestamp for \textbf{(a)} model trained with randomly sampled timestamp manipulations and \textbf{(b)} model fine-tuned for 10 additional epochs sampling manipulations close to the ground-truth timestamp. Each curve represents a month shift ($\Delta t_{\textit{month}}$), while the x-axis denotes different hour shifts ($\Delta t_{\textit{hour}}$). By fine-tuning it with harder-to-detect tampering, the model learns to identify subtler manipulations of a few hours and months (bottom left region of the plots), improving the detection rate of such cases.}
            \label{fig:range_tampering}
        \end{figure*}

    \subsection{Sensitivity Analysis: Timestamp Manipulation}\label{sec:sensitivity_timestamp}
        In real scenarios, when the timestamp of a photograph is modified, the new time-of-capture is typically selected considering a plausible, often close to the original, moment in time as a way to make the detection harder. For example, claiming a $9$\textsc{AM} image was captured in daylight hours is more convincing than saying it was captured at night.
        
        To evaluate this scenario, we performed several experiments in which all images in our test set were tampered with by an equal month and hour shift in both directions to its original time-of-capture. For a given shift pair ($\Delta t_{\textit{month}}, \Delta t_{\textit{hour}}$), we tampered with every testing image by adding $\pm\Delta t_{\textit{month}}$ and $\pm\Delta t_{\textit{hour}}$ to its time of capture, effectively producing four tampered versions for each image. For example, for $\Delta t_{\textit{month}} = 1$ and $\Delta t_{\textit{hour}} = 2$, a picture captured in (Dec, $11$\textsc{AM}) would generate timestamps (Jan, $9$\textsc{AM}), (Jan, $1$\textsc{PM}), (Nov, $9$\textsc{AM}) and (Nov, $1$\textsc{PM}). We evaluated the detection rate of such manipulated images and present the results in Figure~\ref{fig:range_tampering_random}. 
    
        As we expected, the closer an alleged timestamp is to the original moment of capture, the harder it is to detect that tampering occurred. Shifts by a single hour or month were detected in less than $20\%$ of the cases. Due to the similarity in the appearance of the scene, these are complex to detect solely based on pixel values and without any other hard assumptions on the images. This becomes even more challenging considering that modern cameras make use of automatic exposure adjustment techniques that compensate for brightness distortions during capture, making a scene darker or brighter than it is in reality.  Besides obliterating image differences caused by 1-hour time spans, artificial changes in the brightness of the picture might be perceived as an alteration caused by the progression of time, thus influencing the model decision. Existing works able to detect such manipulations more accurately either consider weather and astronomical information~\cite{ghosh2017detection} or rely on several combined images of the same time and place~\cite{chen2019deep}, which are not always available for images originated from social media.
        
        As the gap between ground truth and alleged timestamp increases, the inconsistency between time and appearance becomes more apparent and detectable. E.g., images that were presented as being captured in a different season ($\Delta t_{month} \geq 3$) or period of the day ($\Delta t_{hour} \geq 6$, such as claiming a morning scene was captured at noon or at night) were detected in more than $75\%$ of the cases.
        
        As we trained our models with random sets of tampered month and hour, in accordance with the evaluation protocol from~\cite{salem2020learning}, the network learns how to better capture clearer manipulations, as they tend to be sampled more frequently in comparison to sets closer to the ground-truth timestamp. However, it is possible to shift the focus to detecting subtler and harder manipulations as well. For that, we fine-tuned our best model (DenseNet — G, t, l, S - TA) for 10 epochs using tampered timestamps closer to the ground-truth time-of-capture. Each manipulated timestamp was generated randomly selecting $\Delta t_{month}$ and $\Delta t_{hour}$ in $\{\pm1, \pm2\}$. We present the results in Figure~\ref{fig:range_tampering_ngb}. The detection rate for such cases improves considerably as the model now captures the contrast between neighboring hours and months. In the general/random case, this model has a minor decrease in accuracy ($77.78\%$) while slightly improving AUC ($0.888$).

        \begin{table}[!t]
            \centering
            \caption{Sensitivity analysis for different degrees of noise ($\Delta l$) in the location coordinates and satellite imagery.} 
            \label{tab:sensitivity_location}
            \resizebox{\linewidth}{!}{%
            \begin{tabular}{@{}rcccccc@{}}
            \toprule
            &  \multicolumn{2}{l}{Proposed Approach} &  \multicolumn{2}{l}{Location Augmentation} & \multicolumn{2}{c}{Salem et al.~\cite{salem2020learning}}\\ \cmidrule(lr){2-3} \cmidrule(lr){4-5} \cmidrule(lr){6-7}
            \multicolumn{1}{c}{$\Delta l$}      &  Acc (\%)      & AUC    &  Acc (\%) & AUC &  Acc (\%) & AUC\\ \midrule
            \multicolumn{1}{c}{Without noise} & $81.1$ & $.885$ & $75.9$ & $.848$ & $59.0$ & $.627$\\\hdashline
            $1^{\circ}$   &  $66.3$  & $.718$ &  $75.9$  & $.848$     &  $52.2$  & $.539$\\
            $5^{\circ}$   &  $66.3$  & $.721$ &  $76.0$  & $.847$     &  $52.1$  & $.538$\\
            $15^{\circ}$  &  $65.9$  & $.716$ &  $75.8$  & $.846$    &  $51.9$  & $.533$\\
            $30^{\circ}$  &  $64.3$  & $.695$ &  $74.8$  & $.838$    &  $51.6$  & $.529$\\
            $45^{\circ}$  &  $62.2$  & $.686$ &  $73.5$  & $.824$    &  $51.6$  & $.528$\\
            $60^{\circ}$  &  $61.1$  & $.677$ &  $72.5$  & $.811$    &  $51.6$  & $.528$\\
            $75^{\circ}$  &  $60.5$  & $.671$ &  $71.7$  & $.801$    &  $51.6$  & $.527$\\\bottomrule         
            \end{tabular}%
            }
        \end{table}

        \begin{figure*}[!t]
            \centering
            \includegraphics[width=0.85\textwidth]{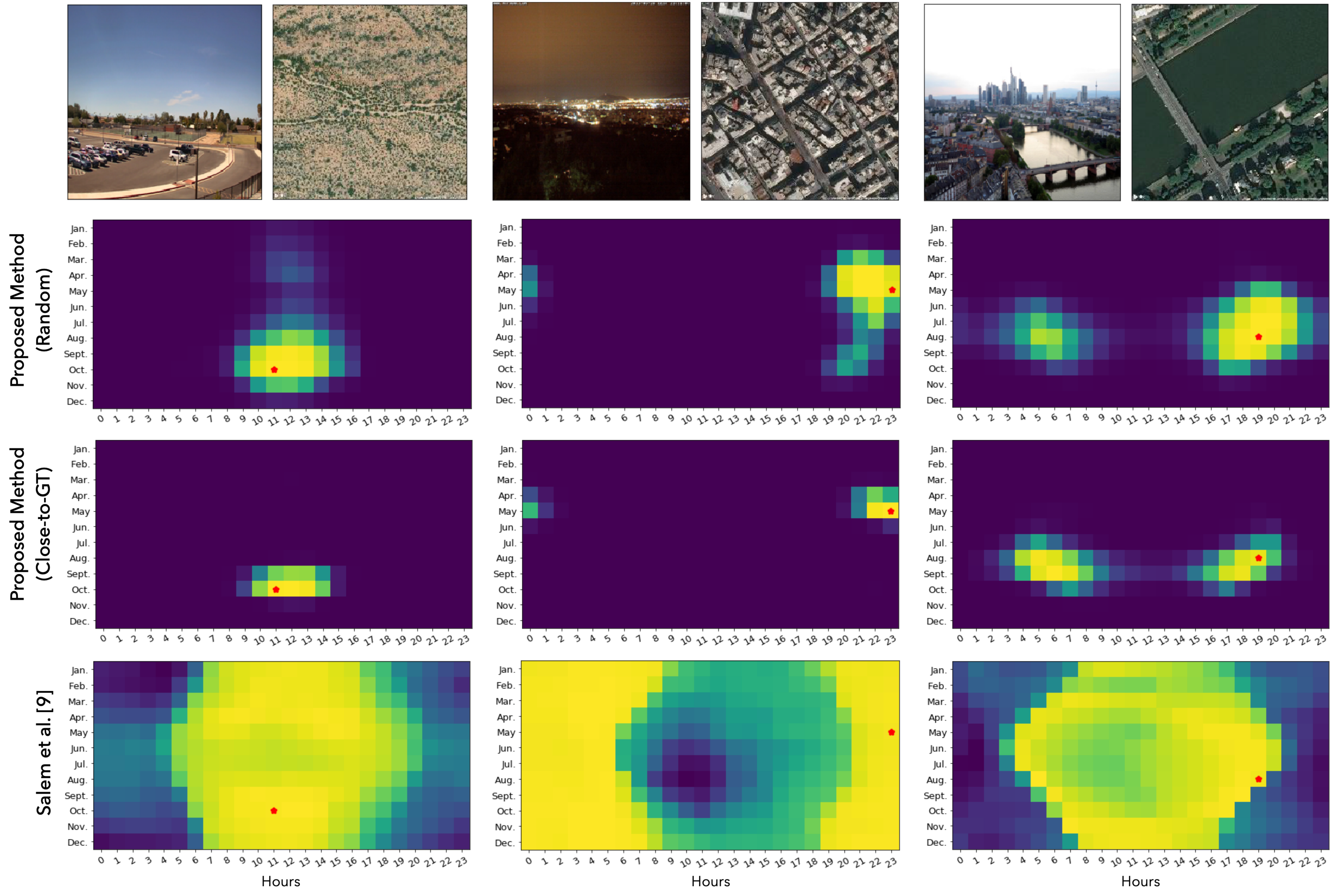}
            \caption{Heatmap of the consistency probability distribution over local time (all possible months and hours of capture) for proposed method trained with randomly sampled manipulations and alleged timestamps close to the ground truth, in comparison to Salem et al.~\cite{salem2020learning}. For each example, we show the ground-level picture and satellite image of that location. Yellow areas represent high consistency, and blue indicates inconsistent moments in time. The red dot marks the ground-truth moment-of-capture. Note how the proposed methods provide high-consistency month/hour pairs close to the ground-truth timestamp with fewer false activation regions in comparison to Salem et al.~\cite{salem2020learning}. We provide additional examples in the Supplementary Material.}
            \label{fig:time_estimation}
        \end{figure*}

    \subsection{Sensitivity Analysis: Geographic Location}\label{sec:sensitivity_location}
        As shown in our ablation study, the geographic coordinates and satellite image provide an essential context to assess the consistency between an alleged timestamp and the ground-level photograph. However, in real cases, the precise location in which an image was captured might not be known, and only a rough set of coordinates might be available instead. In this sense, we evaluate the impact of noisy location coordinates on the performance of our approach. For each sample in our testing set, we add $\pm\Delta l$ to its original latitude or longitude, artificially altering the location of that image. As the satellite image is directly related to the geographic coordinates, we also collect a new aerial photograph for the perturbed position. We evaluate the accuracy and AUC under different values of $\Delta l$ for our best ablation model and the method of Salem et al.~\cite{salem2020learning}, and present the results in Table~\ref{tab:sensitivity_location}. 
        
        Feeding noisy sets of coordinates and satellite photograph to the models, as expected, negatively impact their performance. This highlights the sensitivity of the models when we assume correct locations but evaluate it under a noisy scenario. The performance of our approach is stable when considering perturbations smaller than $15^{\circ}$. Each longitudinal movement of $15^{\circ}$ represents a one-hour shift, which is the smallest manipulation captured by the method. In this sense, minor degrees of noise do not affect the performance of the approach. However, when the noisy coordinates deviate considerably from the ground-truth location, the model interprets these differences as signs of manipulation. Most error cases are related to consistent pairs of ground-level image and timestamp that are mistakenly classified as manipulations due to inconsistent coordinates. It also might indicate that the model is over-reliant on the coupled information of ground-level picture and geographic input data, and visual inconsistencies that might be created by combining a photograph with a satellite image from another location (e.g., an urban scene paired with a rural or coastal aerial picture) are identified as a sign of tampering. This is especially true for samples from the Flickr subset of CVT, as each geographic location might be represented by a single picture. We present in the Supplementary Material an extension of this experiment, evaluating more degrees of noise and discriminating the results per subset of the CVT dataset.
        
        Considering this, to improve robustness to unreliable location information, we retrain our model with randomly perturbed geographic coordinates. In each training batch, we jitter the location coordinates of $50\%$ of batch samples by a noise $\pm\Delta l \in \{0.05^{\circ}, 0.1^{\circ}, 0,25^{\circ}, 0.5^{\circ}, 1^{\circ}, 5^{\circ}, 10^{\circ}, 15^{\circ}\}$. The training is performed as detailed in Section~\ref{sec:dataset_training_details} and we include the results in Table~\ref{tab:sensitivity_location}.
        
        The \textit{Location Augmentation} strategy outperformed the other methods when we consider noisy geographic information. The impact of unreliable location on it is less noticeable than that observed in our model optimized with ground-truth geographic information and the approach from Salem et al~\cite{salem2020learning}. This highlights the importance of introducing techniques that improve the robustness to location errors, especially if we want to apply these models in unconstrained scenarios, such as when verifying footage from social media.

    \subsection{Time Estimation: A Qualitative Exploration}\label{sec:time_estimation}
        We also investigated the application of our method when a timestamp is not available. For a ground-level photo with associated location and satellite image, we predicted the consistency probability $P(y | G, t_{i}, l, S)$, $\forall$ $t_{i} \in T$, with $T$ being all combinations of month and hour of capture. 
        
        In Figure~\ref{fig:time_estimation}, we show the consistency distribution heatmap for a few examples, as well as their ground-truth timestamp. We also include the consistency heatmaps produced by our model fine-tuned to detect subtler manipulations (Section~\ref{sec:sensitivity_timestamp}). In comparison, we generate heatmaps with the method from Salem et al.~\cite{salem2020learning}, computing the consistency score for all possible timestamps in a similar manner to Section~\ref{sec:exp_ablation}. 
        
        Even though our method is not explicitly trained to predict time, it is able to coherently estimate a possible span in which an image might have been captured. As it properly learned the influence of time in the appearance of a scene, capturing intrinsic temporal patterns, it also produces more precise estimations than~\cite{salem2020learning}. The model optimized to detect subtler manipulations is even better at assigning areas of high consistency closer to the ground-truth timestamp with less false activation on other month/hour pairs.

    \subsection{Summer Snow and Midnight Sun: Tampering Telltales}\label{sec:xai}
\begin{figure}[!t]
            \centering
            \includegraphics[width=0.95\linewidth]{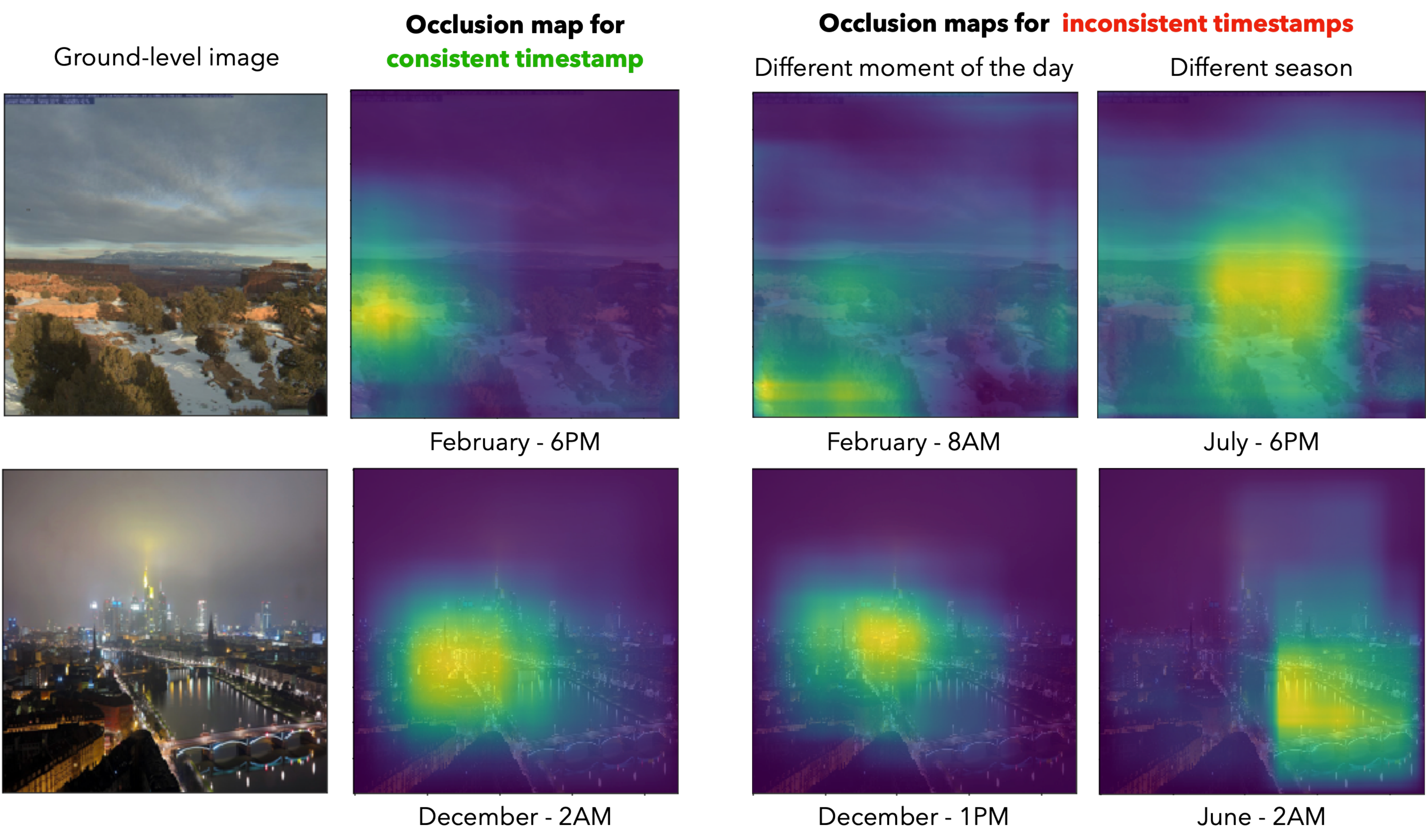}
            \caption{Occlusion activation maps~\cite{zeiler2014visualizing} for two pictures under different timestamps. The network focuses on specific elements depending on the alleged timestamp, with yellow regions representing important elements and blue areas having a low impact on the decision. Activation tends to be spread homogeneously throughout the image, as the network focuses on the illumination and overall appearance of the scene. However, the model is also able to shift its attention to clear telltales, such as city lights during the day. We provide additional examples in the Supplementary Material.}
            \label{fig:xai_occlusion}
        \end{figure}

        As humans, we have a good intuition of which elements in a scene might be inconsistent with an alleged time of capture. For example, we can easily spot the inconsistency of a bright sky in a nighttime picture or of snow-covered ground in a summer scene. Besides achieving high accuracy in tampering detection, we want to provide direct explanations of what elements in the scene might be the cause of inconsistency according to our models.  
        To this end, we investigated which image regions have the most impact on the decision of the network for both consistent and inconsistent examples. Similarly to~\cite{zeiler2014visualizing}, we occluded, in a sliding window manner, parts of the input image with gray patches of sizes $50\times50$, $100\times100$, $100\times50$ and $50\times100$ and evaluated the difference in consistency probability $y$. When an important area of a consistent image is occluded, we expect $y$ to decrease; whereas, occlusions over an inconsistent sample might increase $y$, as it might hide discrepant elements. Figure~\ref{fig:xai_occlusion} depicts the occlusion maps for two ground-level images, considering the ground-truth and manipulated timestamps. Additional examples are presented in the Supplementary Material.
        
        The network focuses on specific elements depending on the time of capture being examined. When claiming the picture was captured in a different moment of the day, the model often considers regions spread across the image, as the scene changes uniformly with the shift in illumination. Besides that, it also activates for some inconsistent elements, such as shadows (bottom-left corner of top example) and nighttime city lights (bottom example). Similarly, when tampering with their month, it considers background vegetation and lights reflected by the water, in the top and bottom scenes, respectively.
        
        Besides the occlusion maps, the transient attributes ($a_{G}$ and $a_{S}$) could provide additional evidence for the network decision. Even though they do not significantly improve the detection performance, as shown in our ablation study (Section~\ref{sec:exp_ablation}), they encode how our model perceived the appearance of the scene based on two distinct sets of inputs. As $a_{G}$ is obtained strictly from the ground-level image and is not influenced by a possibly manipulated timestamp, it captures the characteristics of the scene when the picture was taken. Differently, $a_{S}$ is predicted from the satellite photo, location coordinates, and the alleged timestamp, predicting the expected scene appearance at that alleged moment. In this sense, in case the timestamp $t$ matches the ground-truth moment-of-capture, we expect $a_{G}$ and $a_{S}$ to be similar, whereas an inconsistent $t$ might lead to discrepancies in both sets of attributes. The analysis of such inconsistencies is useful for explainability purposes.
        
\begin{figure}[!t]
            \centering
            \includegraphics[width=\linewidth]{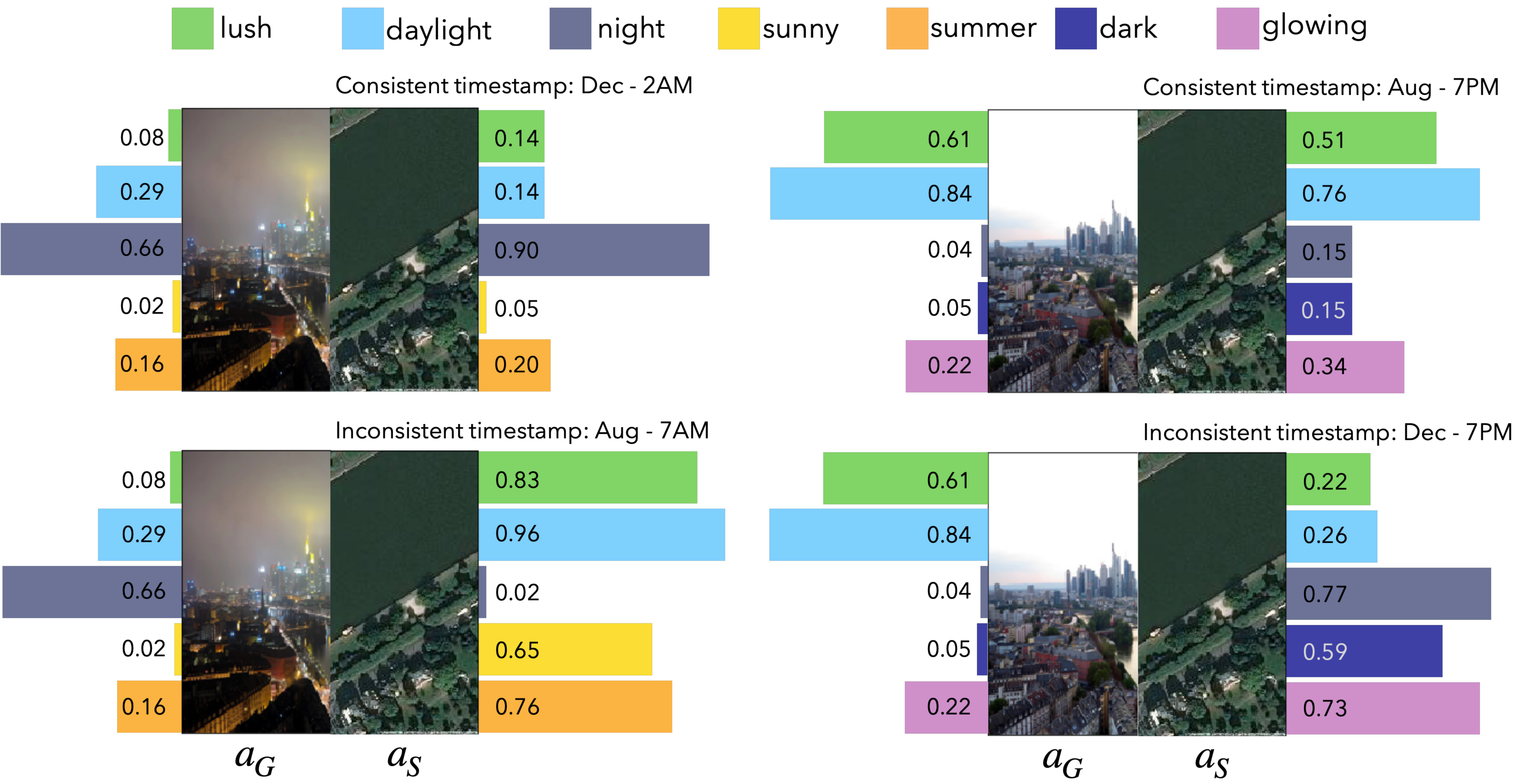}
            \caption{Comparison between a subset of transient attributes $a_{G}$ and $a_{S}$ for two scenes, for consistent (top) and inconsistent (bottom) timestamps. Our model correctly matches both sets of attributes for consistent timestamps. However, as it estimates $a_{S}$ without using ground-level signal, these attributes substantially differ from $a_{G}$ for inconsistent time-of-captures. We provide additional examples in the Supplementary Material.}
            \label{fig:xai_transient}
        \end{figure}

        We compare the sets of transient attributes for the same location taken at different moments. For each scene, we extracted attributes $a_{S}$ considering two timestamps, one matching the ground-truth time-of-capture and another from a tampered timestamp. We present in Figure~\ref{fig:xai_transient} their comparison against $a_{G}$. To highlight their differences, we selected the top five divergent attributes from the inconsistent setup.
        
        When considering the same moment in time, $a_G$ and $a_S$ tend to be similar, while an inconsistent timestamp will produce discrepant transient attributes. For the inconsistent example on the left column of Figure~\ref{fig:xai_transient}, our model expected it to be a sunny day of summer for the (August, $7$\textsc{AM}) time-of-capture, in contrast to the nighttime scene in the ground-level image (December, $2$\textsc{AM}). We provide more examples in the Supplementary Material.

    \subsection{Transient Attribute Influence Analysis}
    
        \begin{figure*}[!t]
            \centering
            \subfigure[Top-ranked attributes.]{
                \includegraphics[width=0.95\columnwidth]{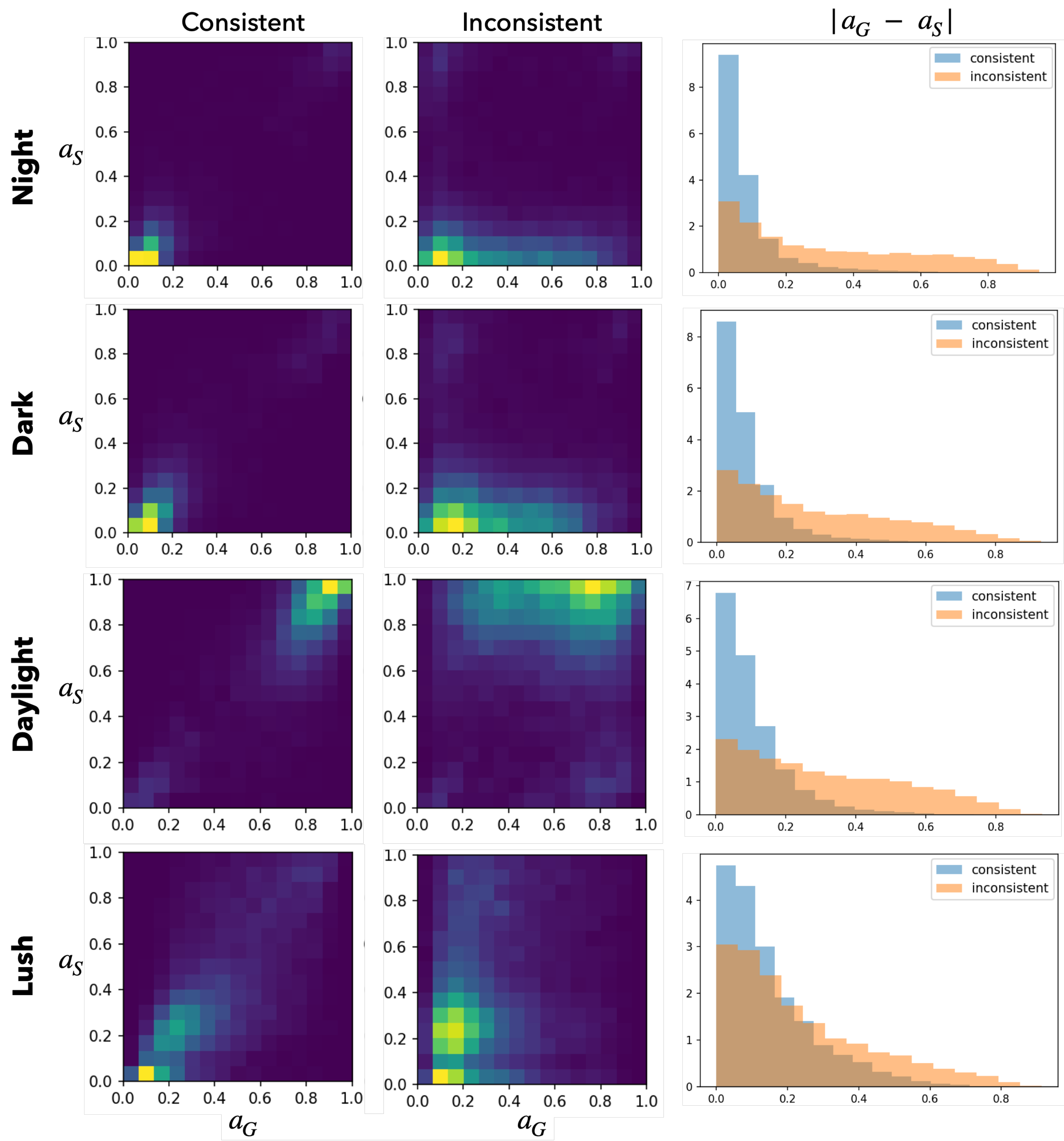}
                \label{fig:ta_hist_top}
            }
            \hfill
             \subfigure[Bottom-ranked attributes.]{
                \includegraphics[width=0.95\columnwidth]{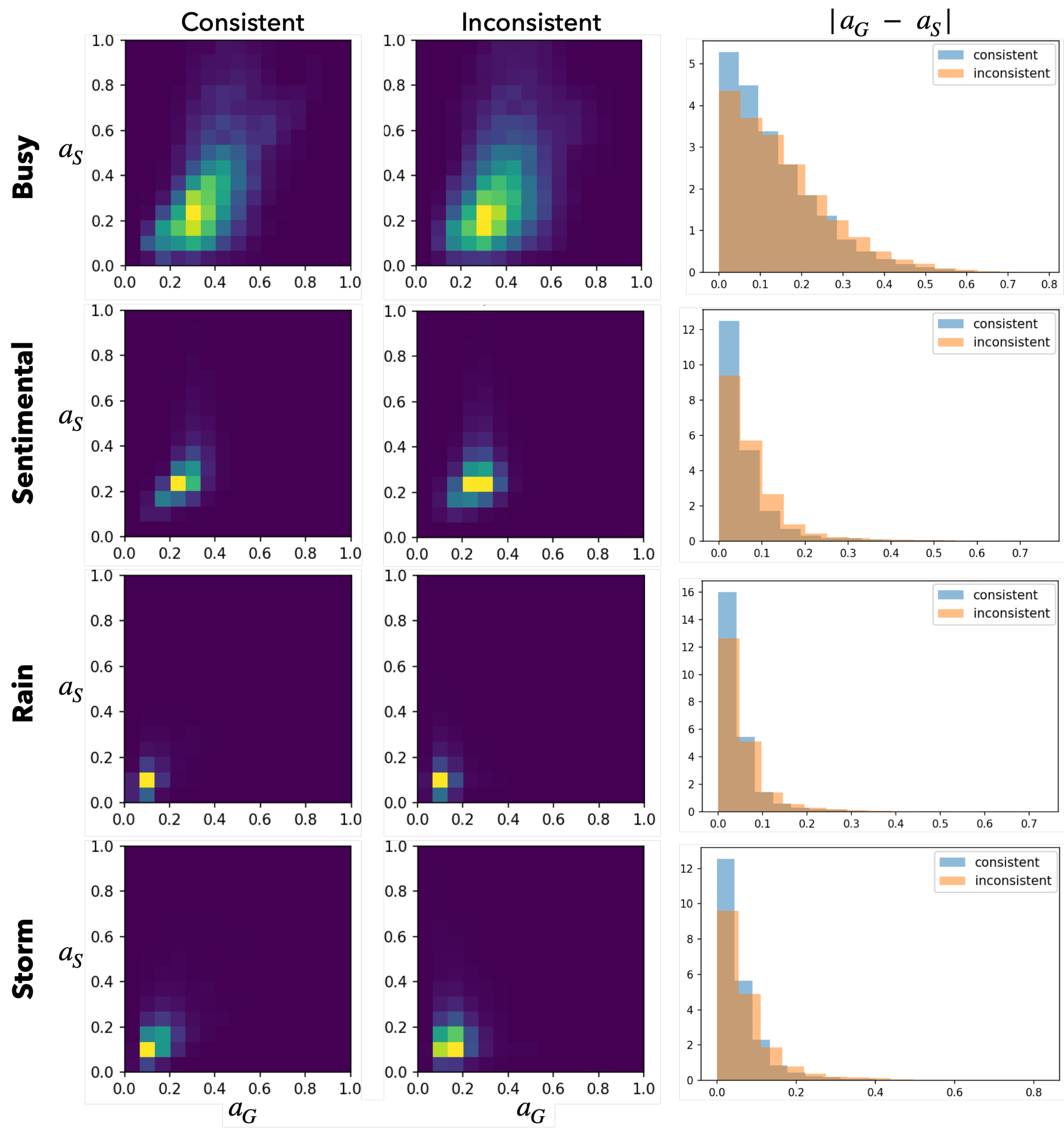}
                \label{fig:ta_hist_bottom}
            }
            \caption{Joint histograms for eight transient attributes estimated from ground-level picture ($a_G$) or satellite image, location and timestamp ($a_S$) for consistent and inconsistent examples in our test set. Attributes were ranked based on the difference in mutual information between the joint histograms of both classes, highlighting \textbf{(a)} discriminative and \textbf{(b)} less informative attributes. We also show the histogram over $|a_G - a_S|$, emphasizing the difference in distributions of each class.}
            \label{fig:ta_hist}
            \vspace{-0.3cm}
        \end{figure*}
    
        The distribution of transient attributes estimated by our network reflects some aspects of scene appearance that might be inconsistent with the alleged timestamp. As shown in Figure~\ref{fig:xai_transient}, most inconsistent attributes are related to illumination and seasonal patterns, which might indicate they are discriminative enough to verify a timestamp.     
        
        Figure~\ref{fig:ta_hist} presents an analysis of how the network decision is reflected into the set of estimated transient attributes. For each transient attribute in $a_G$ and $a_S$, we computed a joint histogram over their predicted values for consistent and inconsistent examples in our testing set that were classified with more than $90\%$ confidence. As expected, estimated attributes are similar for consistent examples, with a high count located in the diagonal of the histograms. Whereas for inconsistent examples, the divergence between $a_G$ and $a_S$ predictions is highlighted by activation spread outside the diagonal.

        For every histogram, we computed the mutual information (MI) considering the distributions obtained by $a_G$ and $a_S$. This measures the dependency between an attribute estimated from the ground-level image or from satellite, location, and timestamp. In this sense, the MI calculated from consistent examples tends to be higher due to the similarity between $a_G$ and $a_S$ than that of inconsistent examples. Considering this, we ranked each transient attribute based on the difference in MI from consistent and inconsistent classes. This quantifies how discriminative an attribute is to capture divergences in $a_G$ and $a_S$ due to a tampered timestamp. Figures~\ref{fig:ta_hist_top} and~\ref{fig:ta_hist_bottom} show the top and bottom four ranked attributes, respectively. 
        
        Most discriminative attributes are related to the day-night cycle and changes in the illumination of the scene, with varied histogram distribution. Complementarily, \textit{lush} captures the vegetation of scenes, an important element across months and seasons. On the other hand, attributes such as \textit{busy} and \textit{sentimental} show similar histograms between both classes, while \textit{rain} and \textit{storm} are concentrated on a few bins, which might reflect the low occurrence of images with such conditions in our dataset. 
        
        We also present, for these eight attributes, the normalized histogram of $|a_G - a_S|$ for both classes. For discriminative attributes, the distributions are considerably different, with consistent examples concentrated on lower bins and inconsistent ones spread towards higher differences between attributes. Differently, less informative attributes show similar histogram distribution for both consistent and inconsistent examples. We present in the Supplementary Material additional experiments comparing the estimated set of attributes for both classes.

\section{Cross-camera evaluation}\label{sec:cross_camera}
    The standard evaluation protocol of Cross-View Time dataset~\cite{salem2020learning} allows for certain cameras to be shared between training and testing sets. This protocol can emulate scenarios in which we need to verify the authenticity of images from a particular set of devices and locations. Considering the ubiquity of surveillance systems (CCTV) nowadays, this is a common scenario, especially for big cities and high visibility events (e.g., protests, musical concerts, terrorist attempts, sports events). In such cases, we can leverage the availability of historical photographs of that device and collect additional images from previous days, months, and years. This would allow the model to better capture the particularities of how time influences the appearance of that specific place, probably leading to a better verification accuracy. However, there might be cases in which data is originated from heterogeneous sources, such as social media. In this sense, it is essential that models are optimized on camera-disjoint sets to avoid learning sensor-specific characteristics that might not generalize accordingly for new imagery during inference.
    
    With this in mind, we propose a novel organization for CVT dataset.\footnote{The data organization of the camera-disjoint sets is available in \href{https://github.com/rafaspadilha/timestampVerificationTIFS}{\repoLink{https://github.com/rafaspadilha/timestampVerificationTIFS}}.} We split available data into training and testing sets, ensuring that all images from a single camera are assigned to the same set. During this division, we aimed to keep the size of each set roughly similar to the original splits, allowing models to be optimized with similar amounts of data. 
    
    Under this cross-camera protocol, we retrained our best model (DenseNet -- $G$, $t$, $l$, $S$, TA), achieving an accuracy and AUC of $67.9\%$ and $.749$, respectively. When we consider the model optimized without the satellite image (DenseNet -- $G$, $t$, $l$), it achieves $65.7\%$ and $.735$ on the same metrics. Even though the aerial image improves performance similarly to our previous experiments, the location-only model is still competitive. In comparison, we trained the method from Salem et al.~\cite{salem2020learning}, using code provided by the authors, and obtained an accuracy of $55.0\%$ and AUC of $.565$. Both proposed models outperformed by a large margin the technique from~\cite{salem2020learning}, highlighting the importance of combining the information from multiple input modalities and training specifically for the timestamp verification task. 
    
    Despite that, we see a performance drop for all methods in comparison to the previous experiments. This was expected due to the more challenging conditions of the new protocol. As there will be locations not covered in the training set, for such cases, the network must take into account in its decision the knowledge obtained from other training samples roughly from that latitude and longitude interval. Besides that, under this new protocol, the models are required to generalize to spatially-distributed sources without relying on camera-specific cues. This is particularly important considering that roughly a third of CVT dataset is originated from 50 static outdoor webcams of the Archive of Many Outdoor Scenes~\cite{jacobs2007consistent}. In a scenario with shared devices between sets, the model could focus on learning patterns specific to these cameras, which allows it to achieve superior accuracy at inference but hinders its generalization to unseen places and devices. 
    
    Nonetheless, as more data becomes available, covering more locations and timestamps, we expect the performance on both camera-disjoint and shared-camera setups to be similar.

\section{Conclusion}
    We introduced a novel approach for detecting if the alleged capture time (hour and month) of an outdoor image has been manipulated. Our architecture incorporates inputs from location, time, and satellite imagery, and is jointly optimized to detect timestamp tampering and estimate high-level attributes about the scene appearance. The proposed approach achieved high detection accuracy, improving the state-of-the-art on a large-scale dataset and surpassing existing approaches while having fewer assumptions and more realistic applicability. 
    
    We demonstrated that incorporating geographical context --- in the form of location coordinates and/or satellite imagery --- was essential for this problem, allowing the model to account for geographic patterns that influence scene appearance in a particular month and hour. Just like location and satellite imagery, our method can be easily adapted to receive other types of data as additional context. High-quality information, such as meteorological data similar to~\cite{ghosh2017detection} or scene classification~\cite{zhou2017places}, could complement present features and improve tampering detection.
    
    While optimizing the model to estimate transient attributes slightly improved the detection accuracy, their major benefit was the added interpretability factor. They allow us to generate explanations for the classification decisions by comparing a set of attributes estimated solely from the ground-level image against another set predicted from the timestamp and location modalities. Mismatching attributes often indicate evidence of tampering captured by our model. Complementarily, we also showed how to interpret an outcome by analyzing pixel-level influence on the model's decisions.
    
    Additionally, we analyzed the sensitivity of our approach under realistic scenarios, with unreliable location information, subtler timestamp manipulation, and changes in the appearance of the scene. Even though most of them influence the model's ability to detect manipulations, their impact can be undermined by applying data augmentation techniques during training. In particular, sampling manipulated timestamps closer to the ground-truth moment-of-capture of images, as well as slightly jittering location coordinates of training samples improved the model robustness to such cases.
    
    We also demonstrated how our method could estimate a possible time of capture for an image missing its timestamp. By shifting how we address the problem, from a verification to an estimation scenario, we show that the proposed method correctly learned discriminative temporal patterns, outperforming existing approaches in this task.
    
    Finally, we propose a new cross-camera evaluation protocol for the Cross-View Time dataset~\cite{salem2020learning}, with camera-disjoint training and testing sets to emulate a realistic application scenario. Under this protocol, trained models are required to generalize for geographically-spread imagery without relying on location- or sensor-specific cues.

\section*{Acknowledgment}
    The authors would like to thank the São Paulo Research Foundation (FAPESP, grants \#2017/21957-2 and \#2019/15822-2) and the US National
Science Foundation (IIS-1553116) for the financial support.


\bibliographystyle{IEEEtran}
\bibliography{strings, refs}

%




\begin{IEEEbiographynophoto}{Rafael Padilha}
is currently pursuing his Ph.D. in Computer Science at the Institute of Computing, University of Campinas, Brazil. Padilha received his M.Sc. in Computing Science in 2017 from the same university. His research interests include machine learning, computer vision, and digital forensics.
\end{IEEEbiographynophoto}

\begin{IEEEbiographynophoto}{Tawfiq Salem}
is a visiting assistant professor in the Department of Computer and Information Technology at Purdue University, USA. He received his Ph.D. in Computer Science from University of Kentucky in 2019. His research interests include machine learning, computer vision, and remote sensing.    
\end{IEEEbiographynophoto}

\begin{IEEEbiographynophoto}{Scott Workman}
received his Ph.D. in Computer Science from University of Kentucky in 2018. His research interests include computer vision and machine learning, specifically focused on understanding the connections between images and their geo-temporal context.
\end{IEEEbiographynophoto}

\begin{IEEEbiographynophoto}{Fernanda A. Andal\'{o}}
is a researcher associated with the Institute of Computing, University of Campinas, Brazil. Andal\'{o} received a Ph.D. in Computer Science from the same university in 2012, during which she was a visiting researcher at Brown University. She is an IEEE member and was the 2016-2017 Chair of the IEEE Women in Engineering (WIE) South Brazil Section. Her research interests include machine learning and computer vision.
\end{IEEEbiographynophoto}

\begin{IEEEbiographynophoto}{Anderson Rocha}
has been an associate professor at the Institute of Computing, the University of Campinas, Brazil, since 2009. Rocha received his Ph.D. in Computer Science from the University of Campinas. His research interests include machine learning, reasoning for complex data, and digital forensics. He was the chair of the IEEE Information Forensics and Security Technical Committee for 2019-2020 term. He is a Senior Member of IEEE.
\end{IEEEbiographynophoto}

\begin{IEEEbiographynophoto}{Nathan Jacobs}
is a professor of Computer Science at the University of Kentucky. He received his Ph.D.\ in Computer Science from Washington University in St. Louis, MO. His research interests include learning-based image understanding, remote sensing, and using crowd-sourced imagery for Earth observation applications. He is a Senior Member of the IEEE.
\end{IEEEbiographynophoto}




\end{document}